\pdfoutput=1
\documentclass[11pt]{article}

\usepackage[]{ACL2023}

\usepackage{times}
\usepackage{latexsym}

\usepackage{graphicx}
\usepackage{enumitem}
\usepackage{caption}
\usepackage{multirow}
\usepackage{array}
\usepackage{subcaption}
\usepackage{xspace}
\usepackage{booktabs}
\usepackage{amsmath}
\usepackage{amssymb}
\usepackage{balance}
\usepackage{fixltx2e}
\usepackage{diagbox}

\usepackage{tabu}% http://ctan.org/pkg/tabu
\usepackage{xcolor}% http://ctan.org/pkg/xcolor
\usepackage{tabularx}% http://ctan.org/pkg/tabularx
\usepackage{booktabs}% http://ctan.org/pkg/booktabs

\newcommand{\ie}{\emph{i.e.,}\xspace}
\newcommand{\eg}{\emph{e.g.,}\xspace}

\newcommand{\say}[1]{``#1''\xspace}
\newcommand{\mname}{OIE-Spec\xspace}

\usepackage[T1]{fontenc}
\usepackage[utf8]{inputenc}

\usepackage{microtype}

\usepackage{inconsolata}
\usepackage[normalem]{ulem}

\title{Shall We Trust All Relational Tuples by Open Information Extraction? \\ A Study on Speculation Detection}

\author{Kuicai Dong$^{1,2}$, Aixin Sun$^1$, Jung-Jae Kim$^2$, Xiaoli Li$^{1,2,3}$ \\
\fontsize{11pt}{12pt}\selectfont
$^1$ School of Computer Science and Engineering, Nanyang Technological University, Singapore\\
\fontsize{11pt}{12pt}\selectfont
\texttt{kuicai001@e.ntu.edu.sg, axsun@ntu.edu.sg}\\
\fontsize{11pt}{12pt}\selectfont
$^2$ Institute for Infocomm Research, A*STAR, Singapore\\
\fontsize{11pt}{12pt}\selectfont
$^3$ A*STAR Centre for Frontier AI Research, Singapore\\
\fontsize{11pt}{12pt}\selectfont
\texttt{\{jjkim, xlli\}@i2r.a-star.edu.sg}}

\begin{document}
\maketitle
\begin{abstract}

Open Information Extraction (OIE) aims to extract factual relational tuples from open-domain sentences. Downstream tasks use the extracted OIE tuples as facts, without examining the certainty of these facts. However, uncertainty/speculation is a common linguistic phenomenon. Existing studies on speculation detection are defined at sentence level, but even if a sentence is determined to be speculative, not all tuples extracted from it may be speculative. In this paper, we propose to study speculations in OIE and aim to determine \textit{whether an extracted tuple is speculative}. We formally define the research problem of \textit{tuple-level speculation detection} and conduct a detailed data analysis on the LSOIE dataset which contains labels for speculative tuples. Lastly, we propose a baseline model \textit{OIE-Spec} for this new research task.\footnote{https://github.com/daviddongkc/OIE\_Spec}

\end{abstract}

%======================================================================
\section{Introduction}\label{sec:intro}
%======================================================================

Open Information Extraction (OIE or OpenIE) aims to generate relational factual tuples from unstructured open-domain text~\cite{yates2007textrunner}. 
The extracted tuples from a sentence are in form of (\textit{ARG}\textsubscript{0}, \textit{Relation}, \textit{ARG}\textsubscript{1}, \dots, \textit{ARG}\textsubscript{n}), also called ``facts'' in OIE. 
By definition,  OIE system is domain-independent, and does not need any input from users. In addition, OIE is highly scalable, thus allowing users to obtain facts with low cost. 
The extracted factual tuples are beneficial to many downstream tasks such as question answering~\cite{khot2017answering}, knowledge base population~\cite{martinez2018openie, gashteovski2020aligning}, and text summarization~\cite{fan-etal-2019-using}. 

The extraction process in OIE mainly depends on sentence syntactical structures (\eg POS tagging, chunking, dependency, constituency, and etc). However, ensuring the factuality/reliability of relational tuples requires accurate semantic understanding of the sentence as well.
% , without much semantic understanding .
Therefore, a key question arises: \textit{Shall we trust all relational tuples extracted by OIE as facts?} Apparently, the answer is no, because uncertainty/speculation is a common linguistic phenomenon, and not all sentences state facts. Downstream applications like knowledge base population typically only require factual tuples, not speculative ones. 

\textit{Speculation} is related to the broader concept of modality which has been extensively studied both in linguistics and philosophy~\cite{sauri2008factuality}. Modality \say{expresses the speaker’s degree of commitment to the events being referred to in a text}~\cite{DBLP:conf/flairs/SauriVP06}.
Other related terms, such as \say{\textit{hedging}}, \say{\textit{evidentiality}}, \say{\textit{uncertainty}}, and \say{\textit{factuality}}, are also used when discussing different aspects of speculation. Extracted information that falls within the scope of speculation cannot be presented as \textit{factual} information~\cite{morante-daelemans-2009-learning}.
Therefore, it is crucial to determine which pieces of extracted information are speculative before using them as ``factual tuples'' in downstream applications. 
However, current studies on OIE do not take speculation into consideration, which can lead to the use of unreliable tuples as facts, affecting the accuracy of various applications.

\begin{table*}
\centering
\resizebox{\linewidth}{!}{%
\begin{tabular}{ l|l|l|l}
 \toprule
 ID& Example sentence &  OIE Tuple with \textbf{speculation} & Meaning\\
 \midrule
  1& Adults \underline{were allowed to} opt out of using computers. & (\textit{adults},  \textbf{can} \textit{opt}, \textit{using computers}) & ability\\
  
  2 & \underline{It is unclear} if the suspects left with any property. & (\textit{suspects}, \textbf{might} \textit{left}, \textit{any property}) & possibility\\
  
3 & The UN \underline{plans to} release a final report in two weeks. & (\textit{the UN}, \textbf{will} \textit{release}, \textit{a final report}) & intention\\
  
  4& Gargling with warm salt water are \underline{reasonable}. & (\textit{warm salt water},  \textbf{should} \textit{gargling}) & suggestion\\
 \bottomrule
\end{tabular}
}
\caption{Examples of speculation annotation in LSOIE dataset. The speculation labels are in boldface as part of tuple relation. To facilitate understanding, we underline the speculation cues in the example sentence, and elaborate the meaning of speculation label under the `Meaning' column. Note that a sentence usually contains multiple tuples, we truncate the long sentence and demonstrate only the tuple with speculation for conciseness.}
\label{tab:spec_example}
\end{table*}

Existing speculation detection tasks are formulated at sentence level~\cite{szarvas-etal-2008-bioscope, konstantinova-etal-2012-review, szarvas-etal-2012-cross, Ghosal-2022-HedgePeer}. The datasets for speculation detection provide sentence-level speculation annotations, and their models are trained to detect the existence of speculation and/or the speculation scope within a given sentence~\cite{adel-schutze-2017-exploring, CHEN2018158, bijl-de-vroe-etal-2021-modality}.
However, OIE systems typically extract multiple tuples from an input sentence. A sentence that is detected as speculative does not necessarily mean that all its tuples are speculative. Additionally, the scopes defined in speculation datasets do not align well with the tuples extracted by OIE. Therefore as a generic task, current sentence-level speculation detection cannot be applied to OIE. 
To bridge this gap, we propose a new research task for OIE that focuses on \textit{tuple-level} speculation detection. In simple words, we expect an OIE model to indicate whether an extracted tuple is speculative or not. 
Such indication of speculation would make the OIE tuples more accurate and reliable for downstream tasks.

To the best of our knowledge, there is no datasets specifically designed for tuple-level speculation detection. Nevertheless, the recently released LSOIE~\cite{solawetz-larson-2021-lsoie} provides a timely preview of this interesting research task. LSOIE is a large-scale OIE dataset converted from the QA-SRL 2.0 dataset~\cite{fitzgerald-etal-2018-large}. We observe that LSOIE provides additional annotation to some OIE tuples with \textit{auxiliary modal words}.  Table~\ref{tab:spec_example} lists a few example sentences and their annotated OIE tuples. A full list of the auxiliary modal words include `\textit{might}', `\textit{can}', `\textit{will}', `\textit{would}', `\textit{should}', and `\textit{had}'; these words express the meaning of possibility, ability, intention, past intention, suggestion, and past event, respectively. It is important to note that \textit{these auxiliary modal words in OIE relations \textbf{may  (43.3\%)}  or \textbf{may not  (56.7\%)} appear in the original given sentences}. In this study, we use these auxiliary modal words as speculation labels for OIE tuples, because they express the degree of certainty at tuple level.

Tuple-level speculation detection is challenging, as it is common for only a certain portion of a sentence to carry speculative semantics. Certain words (\eg \say{\textit{may}}, \say{\textit{if}}, \say{\textit{plan to}}), also known as speculation cues, are responsible for semantic uncertainty, making part of a sentence (or the corresponding extracted tuples) vague, ambiguous, or misleading. In this paper, we develop a simple yet effective baseline model to detect \textbf{O}pen \textbf{I}nformation \textbf{E}xtraction's \textbf{Spec}ulation (called \mname).  \mname detects speculation from two perspectives: \textit{semantic} and \textit{syntactic}.  To model semantics, \mname adds additional relation embedding into BERT transformers, and uses BERT's hidden state of the tuple relation token as \textit{semantic representation}. For syntactic modeling, \mname explicitly models the sub-dependency-graph, \ie immediate neighbours of the tuple relation token in dependency parsing. It adaptively aggregates nodes in the sub-graph using a novel relation-based GCN, and uses the aggregated representation as \textit{syntactic representation}. The concatenated semantic and syntactic representations are then used for speculation detection. 

Our contributions in this paper are threefold. First, we propose a new research task in OIE to detect \textit{tuple-level} speculation. This task examines the reliability of extracted tuples, which aligns well with the goal of OIE to extract only \textit{factual} information.  Second, we conduct a detailed analysis on speculation in LSOIE from two aspects: (i) their presence in language, and (ii) the level of detection difficulty. Third, we propose \mname, a baseline model to detect tuple-level speculation in OIE. \mname leverages both semantic (BERT) and syntactic (Sub-Dependency-Graph) representations. We perform extensive experiments to analyze the research task of tuple-level speculation, and our results show that \mname is effective.

%======================================================================
\section{Speculation Analysis on LSOIE}
%======================================================================

In this section, we review the annotation processes of the QA-SRL Bank 2.0 and LSOIE datasets, with a key focus on speculation. We then study the distribution of speculation labels, by the perceived level of difficulty in speculation detection.

%========================================================
\subsection{Annotation} \label{ssec:lsoie_annotation}
%========================================================

The QA-SRL Bank 2.0 dataset consists of a set of question-answer pairs for modeling verbal predicate-argument structure in sentence~\cite{fitzgerald-etal-2018-large}.\footnote{Question-Answer Driven Semantic Role Labeling.  \url{https://dada.cs.washington.edu/qasrl/}} A number of questions are crowdsourced for each verbal predicate in a sentence, and each answer corresponds to a contiguous token span in the sentence. Examples of QA-SRL annotations can be found in Appendix~\ref{appendix:qa-srl}. Crowdworkers are required to define questions following a 7-slot template, \ie Wh, Aux, Subj, Verb, Obj, Prep, Misc. Among them, `Aux' refers to auxiliary verbs\footnote{There are three types of main auxiliary verbs: `\textit{be}', `\textit{do}', and `\textit{have}'. Besides them, there’s also a special type that affects grammatical mood, called \textit{modal auxiliary verbs}.}, and answers to questions with modal auxiliary verbs reflect speculation information of the fact well. Note that, as all of the questions are crowdsourced, the use of auxiliary modal verbs in questions is based on crowdworkers' understanding of the sentences. The modal verbs may or may not appear in the original sentence.

\begin{table}[t]
\centering
\resizebox{\linewidth}{!}{%
\begin{tabular}{ l|rrrr}
 \toprule
 Subset & \#Sent & \#Tuple & \#Spec. Tuple & \%Spec\\
 \midrule
 wiki\textsubscript{test} &  4,670 & 10,635 & 1,015 & 9.5\%\\
 wiki\textsubscript{train} &  19,630 & 45,931 & 4,110 & 8.9\%\\
 sci\textsubscript{test} &  6,669 & 11,403 & 1,569  & 13.8\%\\
 sci\textsubscript{train} &  19,193 & 33,197 & 4,337 & 13.1\%\\
 \midrule
 Total & 50,162 & 101,166 & 11,031 & 10.9\% \\
 \bottomrule
\end{tabular}
}
\caption{Number of sentences, tuples, tuples with speculation, and the percent of tuples with speculation.}
\label{tab:lsoie}
\end{table}

The QA-SRL Bank 2.0 dataset is then converted to a large-scale OIE dataset (LSOIE) by \citet{solawetz-larson-2021-lsoie}. LSOIE defines $n$-ary tuples, in the form of (\textit{ARG}\textsubscript{0}, \textit{Relation}, \textit{ARG}\textsubscript{1}, \dots, \textit{ARG}\textsubscript{n}) in two domains, \ie Wikipedia and Science. During the conversion, the auxiliary modal verbs in the QA-SRL questions are retained in tuple relations, as shown in Table~\ref{tab:spec_example}. In this study, we consider these  auxiliary modal verbs to reflect speculation. Consequently, in the LSOIE dataset, \textbf{tuples with speculation} are those whose relation contains any of the following six auxiliary verbs: `\textit{might}', `\textit{can}', `\textit{will}', `\textit{would}', `\textit{should}', and `\textit{had}'.

Table~\ref{tab:lsoie} reports the statistics of  sentences, tuples, and the tuples with speculation in the LSOIE dataset.\footnote{We notice that some sentences in the Wiki subset appear again (or repeated) in the Sci subset. In this study, we remove the repeated sentences from the Sci subset. Therefore, our reported numbers differ from that in the original paper.} Overall, 10.9\% of the ground truth tuples contain speculation, indicating the substantial presence of speculative facts in OIE. However, as no OIE system considers speculation, a considerable number of unreliable facts are being extracted.

To the best of our understanding, neither the crowdsourcing process of QA-SRL Bank 2.0 nor the conversion of LOSIE specifically focuses on speculation, as it is not the primary focus of these two datasets. In general, speculation refers to a claim of the possible existence of a thing, where neither its existence nor its non-existence is known for sure \cite{vincze-2010-speculation}. In this work, we follow \citet{matt-2022} to interpret the 6 types of speculation labels as follows:
`\textit{can}' is used to show or infer general ability;
`\textit{will}' and `\textit{would}' are used to show intention or to indicate certainty;
`\textit{might}' is used to show possibility;
`\textit{should}' is used to suggest or provide advice;
`\textit{had}' refers to the past actions or events.

\begin{table}
\centering
\resizebox{\linewidth}{!}{%
\begin{tabular}{ l|r}
 \toprule
 Example sentence & Category\\
 \midrule
 The UN \underline{will release}  a report. & Easy\\
 The UN \underline{will recently release} a report. & Med\\
 The UN \underline{plans to release} a report. & Hard\\
 \bottomrule
\end{tabular}
}
\caption{Examples of 3 cases of speculation according to detection difficulty. All three sentences convey the same fact: (\textit{the UN}, \textbf{will} \textit{release}, \textit{a report}).}
\label{tab:spec_category}
\end{table}

%========================================================
\subsection{Perceived Level of Detection Difficulty} \label{ssec:difficulty}
%========================================================
By analyzing the speculation examples in LSOIE, we have grouped the speculation tuples into three categories, as listed in Table~\ref{tab:spec_category}, based on the perceived level of detection difficulty. (i) The \textbf{easy} category refers to cases where the speculation label (\eg ``\textit{will}'') literally appears in the sentence and is located directly beside the tuple relation (\eg ``\textit{will release}''). (ii) In \textbf{medium} category, the speculation label is present in the sentence, but it does not located directly beside the tuple relation (\eg ``\textit{will recently release}''). (iii) The \textbf{hard} category refers to the cases where the speculation label is not present in the sentence at all (\eg ``\textit{plans to}'' means ``\textit{will}'').

Table~\ref{tab:category_stats} illustrates the distribution of speculation labels across the three levels of difficulty. As seen, 56.7\% of speculative tuples fall under the hard category. It is evident that detecting these hard speculation cases of speculation requires a deep understanding of the semantics of both the input sentence and the corresponding fact. 

Additionally, Table~\ref{tab:spec_label_category} presents a more fine-grained breakdown of the difficulty distribution of speculation labels for each label.
It shows that 92.5\% of `\textit{might}' labels and 81.5\% of `\textit{should}' labels belong to hard category, while the distribution of other labels is more even.

\begin{table}[t] 
% \small
\centering
\resizebox{\linewidth}{!}{%
  \begin{tabular}{l|r|ccc}
    \toprule
     Subset   & \#Spec. Tuple & Easy & Med & Hard \\
    \midrule
    wiki\textsubscript{test} & 1,015 &  20.0\%  & 23.6\%  & 56.4\% \\
    wiki\textsubscript{train} & 4,110  & 20.0\%  & 22.3\%  & 57.6\% \\
    sci\textsubscript{test} & 1,569  & 22.2\% & 19.1\% & 58.8\% \\
    sci\textsubscript{train} & 4,337  & 22.7\%  & 22.4\%  & 54.9\% \\
    \midrule
    Total & 11,031 & 21.3\% & 22.0\% & 56.7\% \\
  \bottomrule
  \end{tabular}
  }
\caption{Distribution of the tuples with speculation by difficulty level.}
\label{tab:category_stats}
\end{table}

\begin{table*}[t] 
\centering
\small
% \resizebox{\linewidth}{!}{%
  \begin{tabular}{l|rr|rr|rr|rr|rr|rr}
    \toprule
     \multirow{2}{*}{\backslashbox{Diff.}{Type}}   & \multicolumn{2}{c|}{can} & \multicolumn{2}{c|}{might} & \multicolumn{2}{c|}{will} & \multicolumn{2}{c|}{should} & \multicolumn{2}{c|}{would} & \multicolumn{2}{c}{had}\\
    & Num & (\%)& Num & (\%)& Num & (\%)& Num & (\%)& Num & (\%)& Num & (\%) \\
    \midrule
    Easy & 1,111 & 36.1 & 114  & 3.9 & 512 & 36.3  & 84 & 8.1 & 349 & 39.7 & 189 & 34.4 \\
    Medium & 1,404 & 33.0 &  104 & 3.6 & 384 & 27.2  & 108 & 10.4 & 267 &  30.3 & 162 & 29.5 \\
    Hard & 1,743 & 40.9 &  2,676 & 92.5 & 516 & 36.5 & 845 & 81.5 & 264 & 30.0 & 198 & 36.1 \\
    \midrule
    Total & 4,258 & 100 & 2,894 & 100 & 1,412 & 100 & 1,037 & 100 & 880 & 100 & 549 & 100 \\
    \bottomrule
  \end{tabular}
  % }
\caption{Number and percentage (\%) of speculation labels by class and by difficulty level.}
\label{tab:spec_label_category}
\end{table*}

%======================================================================
\section{Task Formulation} \label{sec:task}
%======================================================================
As previously discussed, there are no existing OIE systems that take speculation into account when extracting tuples. To fully leverage the capabilities of existing OIE models, it is more meaningful and practical to formulate speculation detection as a \textit{post-processing} task, where the goal is to determine whether a tuple extracted from a sentence is speculative.

Formally, the inputs are the source sentence $s=[w_1, \dots, w_n]$, and a set of relational tuples $T=\{t_1, \dots, t_m\}$ extracted from this sentence\footnote{Each tuple $t_i$ is represented by its components $t_i=[x_1, \dots, x_l]$ where one $x$ is the relation and the rest $x$'s are arguments. Each $x$ corresponds to a contiguous span of words $[w_j, \dots, w_{j+k}]$.}.
The task of \textit{tuple-level speculation detection} is to predict whether a tuple $t_i$ is speculative (\ie a binary classification task), based on $t_i$ and its source sentence $s$.
Note that, the problem can be extended to predict the specific type of speculation defined by six auxiliary modal verbs: `\textit{might}', `\textit{can}', `\textit{will}', `\textit{would}', `\textit{should}', and `\textit{had}'.

%======================================================================
\section{Method: \mname}
%======================================================================

\begin{figure}
    \centering
    %\boxed{
    \includegraphics[trim={0.4cm 0.2cm 0.3cm 0.4cm}, clip, width=0.85\columnwidth]{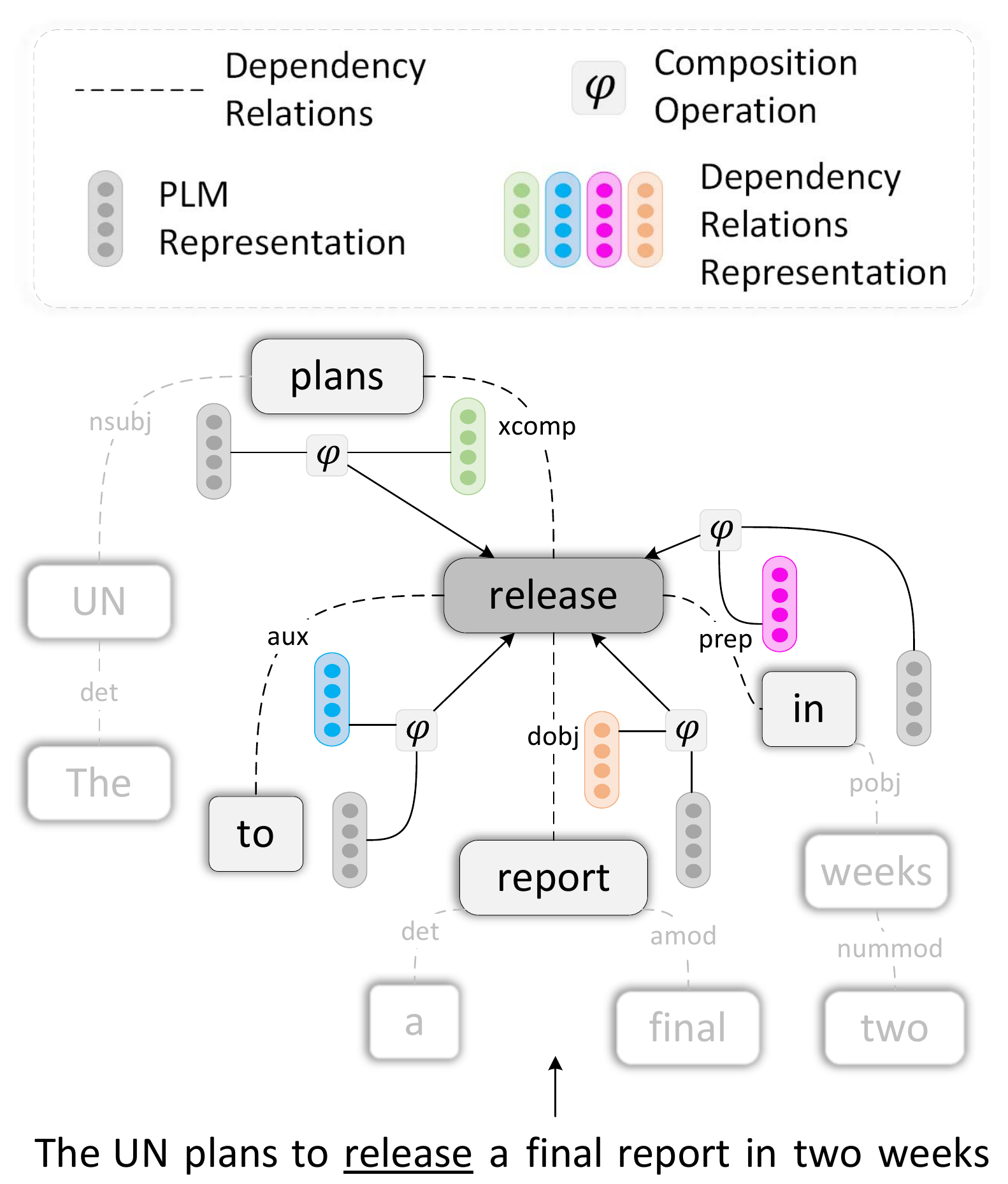}
    %}
    \caption{The relation-based aggregation of \mname. The tuple relation `release' adaptively aggregates its neighbours based on different dependency relations.
    }
    \label{fig:model}
\end{figure}

Given an example tuple (\textit{the UN}, \textit{release}, \textit{a report}), the words ``\textit{plans to}'' in its source sentence which modifies the relation word ``\textit{release}'' strongly indicate speculation. Hence, \mname mainly focuses on the words related to the tuple's relation word in the source sentence for speculation detection. We model these related words using dependency parsing, as shown in Figure~\ref{fig:model}.

%======================================================================
\subsection{Semantic Representation}
%======================================================================
We leverage BERT~\cite{devlin2018bert} as token encoder to obtain token representations.
Specifically, we project all words $[w_1, \dots, w_n]$ into embedding space by summing their word embedding\footnote{If a word maps to multiple sub-words by BERT tokenization, we average the sub-word representations.} and tuple relation embedding:
\begin{equation}
    h_i=\boldsymbol{W_{word}}(w_i)+\boldsymbol{W_{rel}}(w_i)
    \label{eqn:withR}
\end{equation}
where $W_{word}$ is trainable and initialized by BERT word embedding. $W_{rel}$ is a trainable tuple relation embedding matrix. 
The relation embedding is to indicate whether the word $w_i$ is a tuple relation. Specifically, $W_{rel}$ initializes all tokens into binary embeddings (positive embedding for relation word, and negative embedding to non-relation words).
It is worth noing that, we only utilize the tuple's relation word for speculation detection and not its arguments. By including relation embedding, the model explicitly emphasizes the difference between relation tokens and non-relation tokens.

Then, we use $h_s=[h_1, \dots, h_n]$ as the input to BERT encoder and utilize BERT's last hidden states as semantic representations:
\begin{equation}\label{eq:h_sem}
    h_i^{sem} = \mathrm{BERT}(h_i)  \in \mathbb{R} ^ {d_h}
\end{equation}

%======================================================================
\subsection{Syntactic Representation} \label{ssec:spec_cue}
%======================================================================

\paragraph{Speculation Cue Modelling.} 
Speculation cue refers to a minimal unit that indicates speculation \eg ``\textit{might}'', ``\textit{plans to}'', ``\textit{subject to}''. However, such cue words are not explicitly labelled in the source sentence. 
We assume that a speculative tuple is the outcome of \textit{a speculation cue in the sentence directly modifying the tuple's relation word}. 
Therefore, we model the words that are related to the tuple's relation word $v$ through the sentence's dependency structure. Specifically, we extract a sub-dependency-graph $N(v)$ from the sentence dependency which consists of the immediate (or one-hop) neighbours of the tuple relation word $v$. In $N(v)$, each node $u$ directly links to $v$ with an associated dependency relation $r$, as shown in Figure~\ref{fig:model}.

\paragraph{Relation-aware GCN.} 
Dependency graph is heterogeneous, requiring the relation word $v$ to not only know its neighbours $u \in N(v)$, but also the relationships between them $(u,r) \in N(v)$.
Inspired by CompGCN~\cite{vashishth2020compositionbased}, we devise a strategy to embed each dependency relation as a vector and aggregate the corresponding neighbouring nodes together. 
The representation of word $v$, denoted by $h_v$, is updated:
\begin{equation}\label{eq:h_syn}
    h_v^{syn} = f\Bigg(\sum_{(u,r) \in N(v)} \varphi ( u,r) \boldsymbol{W_r} h_u^{sem} \Bigg)
\end{equation}
where $f(\cdot)$ is the activation function, $\boldsymbol{W_r} \in \mathbb{R} ^ {d_h \times d_h}$ is a trainable transformation matrix. $\varphi ( u,r) $ is the neighbour connecting strength, computed based on dependency type:
\begin{equation}
    \varphi ( u,r)  = h_u^{sem} \cdot \boldsymbol{W_{dep}} (r)
\end{equation}
 where $\cdot$ is the dot production operator. $\boldsymbol{W_{dep}} \in \mathbb{R} ^ {d_h \times N_{dep}}$ is a trainable matrix which projects all  dependency relations to the same embedding space as the node $h_u^{sem}$. $N_{dep}$ is the number of unique dependency relations. 

Finally, we concatenate the semantic representation in Equation~(\ref{eq:h_sem}) and the syntactic representation from relation-based GCN in Equation~(\ref{eq:h_syn}) as follows:
\begin{equation} 
    h_v^{final} =  h_v^{sem} \oplus h_v^{syn}
\end{equation}
where $h_v^{final}$ is used by the classification layer to perform binary speculation detection.

As a  binary classification task, we use binary cross-entropy loss:
\begin{equation}
    L_{CE} = - \frac{1}{N} \sum^{N}_{i=1} y_i \mathrm{log}(p_i) + (1-y_i) \mathrm{log}(1-p_i)
\end{equation}
where $N$ is the number of training instances. $y_i$ is the gold standard label, and $p_i$ is the Softmax probability for the $i^{th}$ training instance.

%======================================================================
\section{Experiments}
%======================================================================

%======================================================================
\subsection{Experiment Setup} \label{ssec:setup}
%======================================================================

\paragraph{LSOIE Dataset.}
As mentioned in Section~\ref{sec:task}, we provide the sentence and its relational tuples for tuple-level speculation detection. To avoid potential bias introduced by any specific OIE model, in our experiments, we use tuples which come from the LSOIE dataset.
For the example sentence ``\textit{The UN plans to release a report}", the tuple would be (\textit{the UN}, \textit{release}, \textit{a report}). Our task is to determine whether this tuple is speculative.  In our experiments, we combine two subsets (Wikipedia and Science) into one big dataset. As the result, we have 38,823 sentences and 79,128 tuples for training, originally from wiki\textsubscript{train} and sci\textsubscript{train} sets. We have 11,339 sentences and 22,038 tuples for testing, from the original wiki\textsubscript{test} and sci\textsubscript{test} sets. 

\paragraph{Evaluation Metrics.}
We use \textit{Precision, Recall, and $F_1$ measures} to evaluate this classification task. We report the results from three perspectives: 

\paragraph{(1) Macro-averaged scores.} This is the unweighted average over the two classes, \ie speculative and non-speculative tuples. Micro-average is not used because nearly 89.1\% of tuples are non-speculative, which would dominate the measures.  
\paragraph{(2) Positive scores.} This is the set of measures computed for the positive (\ie speculative) class.
\paragraph{(3) Recall by difficulty.} This measure is the recall of the speculative tuples by perceived difficulty levels (as described in Section~\ref{ssec:difficulty}): the percentage of easy, medium, and hard cases that are correctly detected.~\footnote{Precision is not applicable here, as the task is not to predict the difficulty level of each tuple.}

\begin{table*}
\centering
\resizebox{\linewidth}{!}{%
\begin{tabular}{l|ccc|ccc|ccr}
 \toprule
 \multirow{2}{*}{Models} &
 \multicolumn{3}{c|}{Macro-averaged} &
 \multicolumn{3}{c|}{Positive} &
 \multicolumn{3}{c}{Recall by difficulty} \\
  & $Pr$ & $Re$ &  $F_1$& $Pr$ & $Re$ &  $F_1$ & Easy & Medium & Hard \\
 \midrule
  \textbf{Keyword Matching} (Dep sub-graph) & & & & & & & & & \\
  Modal Verbs & 79.6 & 66.6 & 70.6 & 67.4 & 35.7 & 46.6 & 99.8 & 59.9 & 3.6\\
  Modal Verbs + Top 10 speculative words & 79.3 & 72.3 & 75.1 & 65.4 & 48.0 & 55.4 & 99.8 & 60.8 & 24.5\\
  Modal Verbs + Top 20 speculative words & 77.8 & 72.2 & 74.6 & 62.4 & 48.4 & 54.5 & 99.8 & 60.8 & 25.3\\
  Modal Verbs + Top 30 speculative words & 77.4 & 72.2 & 74.4 & 61.6 & 48.5 & 54.3 & 99.8 & 60.8 & 25.5\\
 \midrule
 \textbf{Semantic-only} (BERT)  & & & & & & & & & \\
 SEM\textsubscript{sentence} & \textbf{86.6} & 72.0 & 76.9 & \textbf{80.0} & 45.6 & 58.1 & 99.8 & 46.4 & 25.6 \\
 SEM\textsubscript{tuple} & 84.0 & 73.2 & 77.1 & 76.0 & 47.9 & 58.8 & 99.8 & 57.6 & 23.2 \\
 SEM\textsubscript{relation}  & 84.5 & 73.1 & 77.3 & 75.7 & 48.3 & 59.0 & 99.8 & 64.4 & 23.4\\
 \midrule
 \textbf{Syntactic-only} (Dependency)  & & & & & & & & & \\
 SYN\textsubscript{sub-graph}& 72.3 & 70.2 & 71.0 & 53.2 & 46.6 & 49.7 & 95.4 & 41.0 & 35.5\\
 SYN\textsubscript{full-graph} & 72.8 & 70.0 & 71.2 & 54.3 & 46.2 & 49.9 & 95.6 & 42.5 & 34.7\\
 \midrule
 \textbf{Semantic} and \textbf{Syntactic} & & & & & & & & &\\
 SEM\textsubscript{sentence} + SYN\textsubscript{full-graph} & 82.4 & 74.0 & 77.8 & 70.5 & 52.3 & 60.1 & 99.8 & 58.0 & 32.8 \\
 SEM\textsubscript{relation} + SYN\textsubscript{full-graph} & 81.1 & 75.4 & 78.3 & 67.2 & 57.0 & 61.7 & \textbf{100} & 62.3 & 37.5 \\
 SEM\textsubscript{relation} + SYN\textsubscript{sub-graph} (\mname) & 80.7 & \textbf{77.5} & \textbf{79.0} & 66.9 & \textbf{58.9} & \textbf{62.6} & \textbf{100}  & \textbf{67.2} & \textbf{40.9} \\

 \bottomrule
\end{tabular}}
\caption{Main results for binary tuple speculation detection. The best results are in boldface.}
\label{tab:baseline_systems}
\end{table*}

%======================================================================
\subsection{Baselines} \label{ssec:baseline}
%======================================================================
As this is a new research task, there are no existing baselines for comparison. We have designed four sets of methods to consider: (i) only semantic representations, (ii) only syntactic representations, (iii) both semantic and syntactic representations, and (iv) keywords matching. 

\paragraph{Semantic-Only.} 
We leverage on BERT to get the semantic representations of the source sentence $s$. Note that all tokens in the sentence are encoded with tuple relation embedding (see Equations~\ref{eqn:withR} and~\ref{eq:h_sem}). As a result, the tuple relation information is explicitly fused to all source tokens, regardless of whether they are in the tuple relation or not. We then evaluate the effectiveness of using weighted pooling of all tokens in sentence (SEM\textsubscript{sentence}), in tuple (SEM\textsubscript{tuple}), and in tuple relation (SEM\textsubscript{relation}) respectively, for speculation detection.

\paragraph{Syntactic-Only.} 
Based on the sentence dependency parsing, we evaluate SYN\textsubscript{full-graph} and SYN\textsubscript{sub-graph}. The former uses a GCN to aggregate node information from the entire dependency graph; while the latter only aggregates the more relevant sub-graph as described in Section~\ref{ssec:spec_cue}.

\paragraph{Semantic and Syntactic.} 
We combine both semantic and syntactic representations and implement two methods. The first one is SEM\textsubscript{sentence} + SYN\textsubscript{full-graph} that uses semantic embedding and dependencies of all tokens in the sentence. In comparison, \mname leverages only the embeddings and sub-graph dependencies of tuple relation token, which is equivalent to SEM\textsubscript{relation} + SYN\textsubscript{sub-graph}.

\paragraph{Keywords Matching in Dependency sub-graph.}
Besides the neural baselines mentioned above, we also experiment with simple keywords matching methods. For these methods, we use a pre-defined speculative keywords dictionary. A tuple is classified as speculative if any of its immediate neighbours in dependency parsing tree contain a word in the dictionary. We first use the 6 modal verbs as the dictionary. Then we include additional frequent speculative keywords (top 10, 20, and 30).\footnote{Speculative keywords selection is in Appendix~\ref{appendix:keywords}.}

%======================================================================
\subsection{Main Results} \label{ssec:results}
%======================================================================
Table~\ref{tab:baseline_systems} reports the experimental results from different baseline methods to be evaluated.
\paragraph{Semantic vs Syntactic.}
The three baselines using semantic representations significantly outperform the two baselines using syntactic representations.  This highlights that the semantic information is more crucial for the model to understand speculation. We also observe that the recall scores of syntactic models are comparable to those of semantic models. By combining both semantic and syntactic information, \mname outperforms all baselines, particularly in recall and $F_1$ scores.

\paragraph{Tuple Relation vs Full Sentence/Tuple.}
With semantic only, the baseline using tuple relation only performs not worse than the two baselines using more information, suggesting that some relations are more likely to be speculative, and the BERT encoding has implicitly considered the contextual information. In terms of syntactic modeling, the baselines using sub-graphs are slightly better than those using full-graphs, indicating that it is valuable to identify speculation cues within the immediate neighbours of the tuple relation.

\paragraph{Neural Methods vs Keywords Matching}
Observe that \mname outperforms all keywords matching methods by a large margin. In particular, the recall of hard case speculations in \mname is nearly double that of keyword matching, indicating the advantages of semantic and syntactic modelling through neural networks.

Overall, \mname achieves the best results. However, the $F_1$ of speculative class is only 62.6, leaving a big room for future investigation.  

%======================================================================
\section{Multi-class Speculation Classification}
%======================================================================

\begin{table}[t]
\centering
\resizebox{\linewidth}{!}{%
\begin{tabular}{ l|rcccr}
 \toprule
  Spec Type & \#N & $Pr$ & $Re$ & $F_1$ & \% Hard\\
 \midrule
 Non-Spec & 19,456 & 93.5 & 97.7 & 95.6 & - \\
 \midrule
 Spec & 2,616 & 66.7 & 46.6  & 54.9 & 57\%\\
 \cmidrule{2-6}
 \quad-- can & 1,003  & 87.6 & 47.9 & 61.9 & 40\%\\
 \quad-- might & 720  & 53.9 & 39.0 & 45.3 & 94\% \\
 \quad-- will & 339  & 82.4 & 59.3 & 69.0 & 38\%\\
 \quad-- should & 283  & 39.4 & 56.2 & 46.3 & 82\%\\
 \quad-- would & 170  & 85.3 & 37.6 & 52.2 & 28\%\\
 \quad-- had  & 101  & 94.4 & 33.7 & 49.6 & 36\%\\
 \bottomrule
\end{tabular}}
\caption{Breakdown results of multi-class speculation classification by speculation class.}
\label{tab:mutli-class}
\end{table}

\begin{table*}[t]
\centering
\resizebox{\linewidth}{!}{%
\begin{tabular}{ l|l|l|l}
 \toprule
 ID& Example Sentence & Spec Cue &  OIE Tuple with Speculation \\
 \midrule
1 & The UN plans to release a final report. & plans to & (\textit{the UN}, \textbf{will} \textit{release}, \textit{a final report}) \\
  
2 & The UN plans to reduce troops. & plans to & (\textit{the UN}, \textbf{might} \textit{reduce}, \textit{troops})\\
3 & Charlemagne planned to continue the tradition. & planned to &  (\textit{Charlemagne}, \textbf{would} \textit{continue}, \textit{the tradition})\\
 \bottomrule
\end{tabular}
}
\caption{Three examples of speculative tuples in LSOIE with similar speculation cue (\say{\textit{plan to}}), but with different speculation labels.
We truncate the long sentence and demonstrate only the tuple with speculation for conciseness.}
\label{tab:spec_ambiguous}
\end{table*}

As mentioned in Section~\ref{sec:task}, we extend the task to predict the both the existence of speculation and its specific type, as defined by the 6 auxiliary modal verbs: `\textit{might}', `\textit{can}', `\textit{will}', `\textit{would}', `\textit{should}', and `\textit{had}'.  
In this way, we perform classification among 7 classes: non-speculative and 6 types of speculative classes.

%======================================================================
\subsection{Results Overview}
%======================================================================
We now extend the binary task to multi-class classification task (\ie 1 non-speculative and 6 speculative classes). Reported in Table~\ref{tab:mutli-class}, the multi-class $F_1$ score decreases to  54.9\%. Detecting speculative type is much more challenging than identifying the existence of speculation as expected.

Table~\ref{tab:mutli-class} also reports the break-down precision, recall, $F_1$ scores of the 6 speculative classes. The performance of these classes are determined mainly by two factors.
(i) The hard cases are naturally more challenging, because the speculation labels do not appear in source sentences. They can only be inferred based on the semantics. We observe that  `\textit{might}' class consists of 94\% of hard cases, leading to the lowest $F_1$ score 45.3\% among all speculative classes. 
(ii) The performance is also affected by the number of instances. The `\textit{had}' class is a small class, with a very low  $F_1$ score 49.6\%, although only 36\% of its labels are hard cases.

%======================================================================
\subsection{Long-tail Distribution}
%======================================================================
\begin{figure}[t]
    \centering
    \includegraphics[width=0.9\columnwidth]{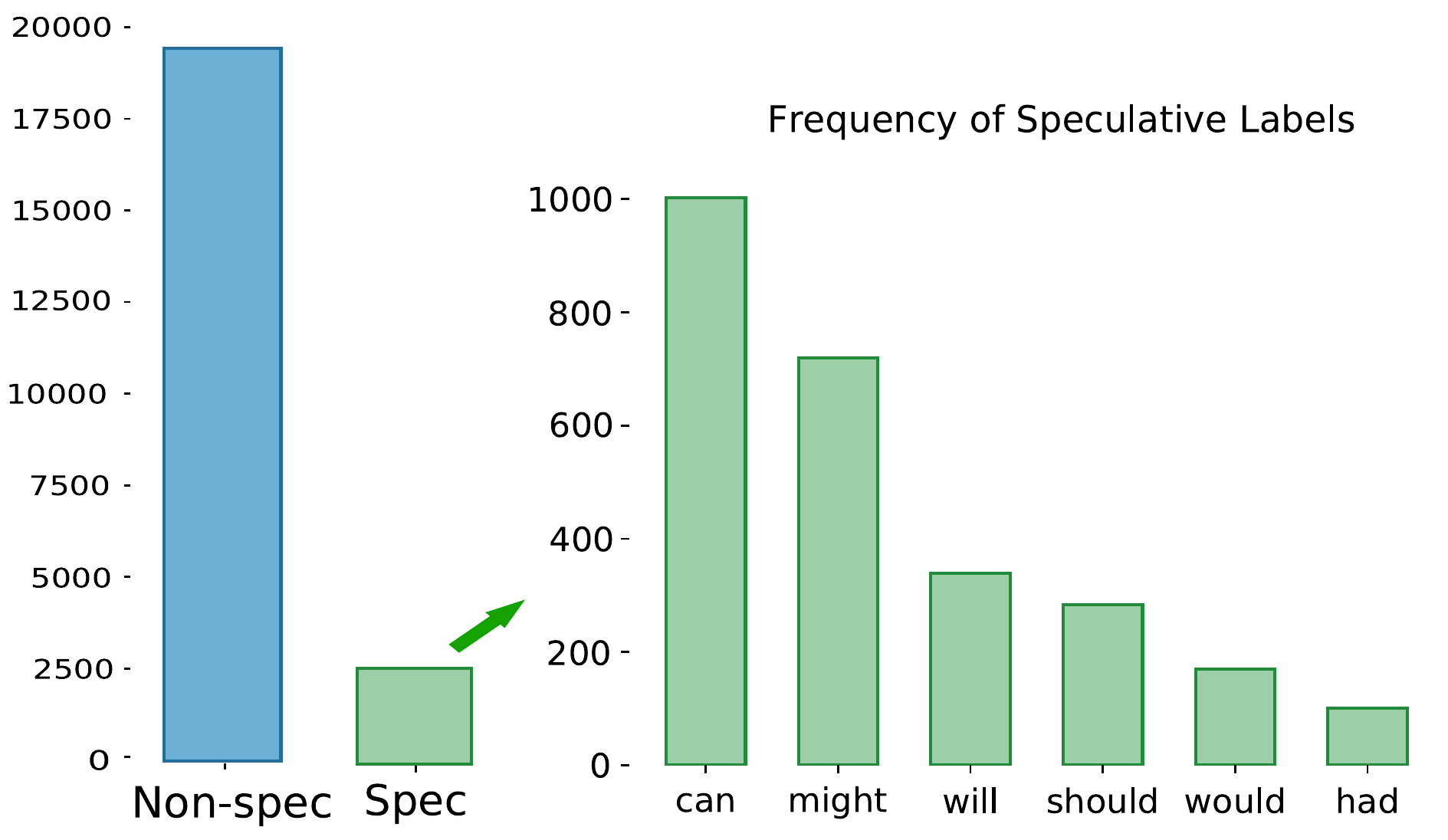}
    \caption{Distribution Non-spec and spec labels, and the break-down distribution of different speculative types.}
    \label{fig:bar}
\end{figure}

The labels exhibit a long-tail class distribution. As shown in Figure~\ref{fig:bar},  10.9\% tuples are speculative. Among them, `\textit{might}' and `\textit{can}' contribute to 65.9\% of the speculative labels. Other speculative types are associated with relatively small number of samples. Such class imbalance is a key source of performance degradation in classification. 
For comparison, the $F_1$ scores of head class `\textit{Non-spec}' and tail class `\textit{Spec}' are 95.6\% and 54.9\%, respectively (with a big difference: 40.7\%).
Among speculative classes, the $F_1$ scores of head class `\textit{can}' and tail class `\textit{had}' are 61.9\% and 49.6\%, respectively (with a big difference: 12.3\%), although the percent of hard cases of `\textit{can}' is higher than that of  `\textit{had}' (40\% > 36\%).

%======================================================================
\subsection{Ambiguous Speculative Types} \label{ssec:ambigous}
%======================================================================

The annotation of speculative types can be ambiguous, due to subtle difference between speculation labels. Table~\ref{tab:spec_ambiguous} lists  three example sentences, all containing \say{\textit{plan to}}, which can express both likelihood and intention. Different annotators might interpret the sentence in different ways, thus annotating the same speculation cue with different labels. Manual checking of annotations in LSOIE shows that such ambiguous labels are rather common, particularly among  hard cases. Such ambiguous hard cases are more challenging to differentiate.

%======================================================================
\section{Related Work}
%======================================================================

\paragraph{Traditional and Neural OIE systems.} Open Information Extraction (OIE) was first proposed by \citet{yates2007textrunner}. 
Before deep learning era, many statistical and rule-based systems have been proposed, including Reverb~\cite{fader2011identifying}, %R2A2~\cite{fader2011identifying}, OLLIE~\cite{schmitz2012open}, 
Clausie~\cite{del2013clausie}, and Stanford OpenIE~\cite{angeli2015leveraging}, to name a few. %OpenIE4~\cite{mausam2016open}, NESTIE~\cite{bhutani2016nested}, and MINIE~\cite{gashteovski2017minie}. 
%We consider them as \textit{traditional} OIE models. 
These models extract relational tuples based on handcrafted rules or statistical methods. The extraction mainly relies on syntactic analysis. %Consequently, any error accumulated in the prior stages deteriorates model performance.

Recently, neural OIE systems have been developed and showed promising results. Neural OIE systems can be roughly divided into two types: generative and tagging-based~\cite{ijcai2022p793}. Generative OIE systems~\cite{cui2018neural, kolluru2020imojie, dong2021docoie} model tuple extraction as a sequence-to-sequence generation task with copying mechanism. 
Tagging-based OIE systems~\cite{stanovsky2018supervised, kolluru2020openie6, dong2022smile_oie} tag each token as a sequence tagging task. Most neural OIE models are designed for end-to-end training and the training data are mostly silver data generated by traditional systems.

\paragraph{Speculation Detection.}
Speculation detection from text is essential in many applications in information retrieval (IR) and information extraction (IE). 
Research in computational linguistics has made significant advances in detecting speculations. Remarkably influential datasets include the BioScope Corpus for uncertainty and negation in biomedical publications~\cite{szarvas-etal-2008-bioscope}, the CoNLL 2010 Shared Task~\cite{farkas-etal-2010-conll} for detecting hedges and their scope in natural language texts, unifying categorizations of semantic uncertainty for cross-genre and cross domain uncertainty detection~\cite{szarvas-etal-2012-cross}, and the SFU Review Corpus~\cite{konstantinova-etal-2012-review} for negation, speculation and their scope  in movie, book, and consumer product reviews.

Current work on speculation is to detect the existence of speculation and/or the speculation cue/scope in a given \textit{sentence}. 
The speculation scope is the maximum number of words affected by the phenomenon.
The speculation cue refers to the minimal unit which expresses speculation.
% The speculation cue refers to the speculative keywords. 
For instance, the most frequent 4 keywords (``\textit{if}'', ``\textit{or}'', ``\textit{can}'', and ``\textit{would}'') contribute to 57\% of the total speculation cues in SFU Review Corpus.
We argue that the detection of such speculation cue at sentence-level is relatively easier. 
In comparison, \textit{tuple-level} speculation cue detection requires not only locating the correct speculative keywords, but also assigning them to the correct tuples. 
It is critical for many real-world applications relying on updated and accurate tuples in knowledge base.

%======================================================================
\section{Conclusion}
%======================================================================
Speculation detection, which is essential in many applications in IR and IE, is not explored in OIE. 
We formally define a new research task to perform \textit{tuple-level} speculation detection on OIE extracted tuples.
We notice that the LSOIE dataset, although not dedicated for speculation detection, provides us a timely preview of this interesting research task.
We conduct a detailed analysis on the speculative tuples of the LSOIE dataset.
To give a glimpse of this research task, we develop a simple but effective model named \mname to detect tuple-level speculation.
As a new research task, we believe there is a big room for further exploration, from dataset construction to more effective models.

\clearpage
%======================================================================
\section*{Limitations}
%======================================================================
We analyze the limitations of our work from three aspects as follows.

\paragraph{Annotation Quality of Speculation.}
Annotating speculation is challenging due to the subtle difference between different speculative types.
As discussed in Section~\ref{ssec:lsoie_annotation}, neither the crowdsourcing of QA-SRL Bank 2.0 nor the conversion of LOSIE pays specific attention to speculation. The annotation of auxiliary modal verbs in the question is based on crowd workers' natural understanding of the sentences. 
Without explicit focus on speculation, the existing annotations reflect the crowd worker's natural understanding without any bias towards specific focusing on speculation.
Therefore, we argue that the annotation is suitable for evaluating tuple-level speculations.

\paragraph{Quality of OIE Tuples.}
We use the ground truth tuples of LSOIE as inputs, to avoid potential bias introduced by any specific OIE model (see Section~\ref{ssec:setup}).
However, existing OIE systems are not so perfect, as we note that the state-of-the-art $F_1$ score in LSOIE dataset is 0.71~\cite{vasilkovsky2022detie}. 
The tuple-level speculation detection, as a post-possessing task, will inevitably suffer from the imperfectly extracted tuples.
However, we argue that \mname can largely mitigate such issue. \mname only relies on the tuple relation for the speculation detection, rather than taking the full tuple with all arguments.
Tuple relation is usually a verb or a verbal phrase that are straightforward to obtain.
Therefore, so long as an OIE system can extract correct tuple relations, \mname can make predictions accordingly.

\paragraph{Modelling Speculation Cues.}
We model the immediate neighbours of the tuple relation in dependency parsing as speculation cues (see Section~\ref{ssec:spec_cue}). However, some speculation cues are not the immediate neighbours of the tuple relation. However, considering the full dependency tree leads to poorer results. 
We leave it for future work to explore a better way to effectively model speculation cues. 

\bibliography{anthology}
\bibliographystyle{acl_natbib}

\appendix

% % \clearpage
%======================================================================
\section{Appendix}
\label{Sec:appendix}
%======================================================================

%=================================
\subsection{Implementation Detail}
\label{appendix:expSetup}
%=================================
% \paragraph{Dependencies.} 
We build and run our system with Pytorch 1.9.0 and AllenNLP 0.9.0 framework. 
The dependency annotations are by spacy-transformers.~\footnote{\url{https://spacy.io/universe/project/spacy-transformers}} There are in total 45 types of dependency labels. 
We tokenize and encode input sentence tokens using bert-base-uncased.~\footnote{\url{https://huggingface.co/bert-base-uncased}} 
The experiments are conducted with Tesla V100 32GB GPU and Intel$^\circledR$  Xeon$^\circledR$ Gold 6148 2.40 GHz CPU.
The experimental results are averaged over 3 runs with different random seeds.
Each epoch is around 15 minutes on a single Tesla V100 32GB GPU.
The hidden dimension $d_h$ for semantic representation $h_i^{sem}$ and that of syntactic representation $h_i^{syn}$ are both 768. 

% \paragraph{Parameters.}

%=================================
\subsection{QA-SRL Annotation}
\label{appendix:qa-srl}
%=================================

QA-SRL stands for Question-Answer (QA) driven Semantic Role Labeling (SRL). It is a  task formulation which uses question-answer pairs to label verbal predicate-argument structure~\cite{fitzgerald-etal-2018-large}. 
Given a sentence $s$ and a verbal predicate $v$ from the sentence, annotators are asked to
produce a set of \textit{wh}-questions that contain $v$ and whose answers are phrases in $s$.
The questions are constrained in a template with seven fields, $s \in$ \textbf{Wh}$\times$\textbf{Aux}$\times$\textbf{Subj}$\times$\textbf{Verb}$\times$\textbf{Obj}$\times$\textbf{Prep}$\times$\textbf{Misc}, each associated with a list of possible options.
Answers are constrained to be a subset of words in the sentence but not necessarily have to be contiguous spans.  
Table~\ref{tab:qa-srl} shows example annotations of an input sentence \say{\textit{The UN plans to release a final report in two weeks.}}. 
In its first annotated question: \say{\textit{When will someone release something}} contains a verb and its answer is a phrase  \say{\textit{in two weeks}} in the sentence.
The answer tells us that \say{\textit{in two weeks}} is an argument of \say{\textit{release}}.
%while the question provides an indirect label on the role that \say{\textit{in two weeks}} plays. 
Enumerating all such pairs provides a relatively complete representation of the verb’s arguments and modifiers.

\begin{table*}[t]
    \centering
    %\resizebox{\linewidth}{!}{%
    \small
    \begin{tabular}{c|ccccccc|c}
    \toprule
         \multirow{2}{*}{Verbal predicate}  & \multicolumn{7}{c|}{7-slot Question} & \multirow{2}{*}{Answer}\\ 
         \cmidrule{2-8}
         & Wh & Aux & Subj & Verb & Obj & Prep & Misc \\ 
         \midrule
         \multirow{3}{*}{release} & When & will & someone & release & something & - & - & in two weeks \\
         & What & will & someone & release & - & - & - & a final report \\
         & Who & will & - & release & something & - & - & The UN \\ 
         \midrule
         \multirow{2}{*}{plans} & Who & - & - & plans & - & to do & something & The UN \\
         & What & does & someone & plan & - & to do & - & release a final report\\
         \bottomrule
    \end{tabular}
    %}
    \caption{The QA-SRL annotations for a newswire sentence: \say{\textit{The UN \textbf{plans} to \textbf{release} a final report in two weeks.}}}
    \label{tab:qa-srl}
\end{table*}

\begin{table*}[t]
    \centering
    \small
    \begin{tabular}{lr|lr|lr|lr}
    \toprule
         \multicolumn{2}{c|}{SFU}  & \multicolumn{2}{c|}{BioScope} & \multicolumn{2}{c|}{FactBank} & \multicolumn{2}{c}{WikiWeasel}\\  
        Cue & Frequency & Cue & Frequency & Cue & Frequency & Cue & Frequency  \\ 
         \midrule
if & 876      & suggest & 810   & expect & 75    & may & 721 \\
or & 820      & may & 744       &  if & 65       & if & 254  \\
can & 765     & indicate & 404  & would & 50     & consider & 250  \\
would & 594     & investigate & 221 & may & 43       & believe & 173 \\
could & 299 &   appear & 213   & could & 29     & would & 136  \\
should & 213 &      or & 197       & possible & 26  & probable & 112 \\
think & 211 &       possible & 185  & whether & 26   & suggest & 108 \\
may & 157      & examine & 183   & believe & 25   & possible & 93 \\
seem & 150    & whether & 169   & likely & 24    & allege & 81 \\
probably & 121     & might & 155     & think & 24     & likely & 80 \\
      - & - &       can & 117      & might & 23     & might & 78  \\
 - & - &  likely & 117    & will & 21      & seem & 67  \\
    - & - &  could & 112     & until & 16     & think & 61  \\
    - & - &  study & 101    & appear & 15    & regard & 58  \\
    - & - &   if & 99         & seem & 11      & could & 55  \\
    - & - &  determine & 87  & potential & 10 & whether & 52  \\
    - & - &   putative & 80   & probable & 10  & perhaps & 51  \\
    - & - &  hypothesis & 77  & suggest & 10   & will & 39   \\
    - & - &   think & 66      & allege & 8     & appear & 32 \\
    - & - &   would & 52      & accuse & 7     & until & 15 \\
         \bottomrule
    \end{tabular}
    \caption{Most frequent speculation cues according to different domains/datasets. The statistics of SFU are from paper~\cite{konstantinova-etal-2012-review} and other statistics are from paper~\cite{szarvas-etal-2012-cross}.}
    \label{tab:spec-cue}
\end{table*}

%=================================
\subsection{Speculative Keywords}
\label{appendix:keywords}
%=================================
% We examine the speculation cues of existing work~\cite{szarvas-etal-2012-cross,konstantinova-etal-2012-review}.

%\paragraph{Re Statistics.} 
Speculation keyword or cue is the minimal unit which expresses speculation~\cite{szarvas-etal-2008-bioscope}.

\citet{konstantinova-etal-2012-review} annotate the Simon Fraser University Review corpus~\footnote{\url{http://www.sfu.ca/~mtaboada/SFU_Review_Corpus.html}} with negation, speculation, and their cues. 
This corpus consists of 400 user reviews from \textit{Epinions.com}.
The 10 most frequent speculative keywords in the SFU Review Corpus are listed in Table~\ref{tab:spec-cue}.

\citet{szarvas-etal-2012-cross} select three corpora (\ie BioScope, WikiWeasel, and FactBank) from different domains, \eg biomedical, encyclopedia, and newswire.
The BioScope corpus~\cite{szarvas-etal-2008-bioscope} contains clinical texts as well as biological texts from full papers and scientific abstracts; the texts are manually annotated for hedge cues and their scopes. The WikiWeasel corpus~\cite{farkas-etal-2010-conll}  is annotated for weasel cues and semantic uncertainty, from randomly selected paragraphs taken from Wikipedia pages. 
The FactBank is a newswire dataset~\cite{DBLP:journals/lre/SauriP09}. Events are annotated in the dataset and they are evaluated on the basis of their factuality from the viewpoint of their sources.
The 20 most frequent speculative keywords in the BioScope, WikiWeasel, and FactBank are also shown in Table~\ref{tab:spec-cue}.

\paragraph{Speculative Keywords Matching.}
As discussed in Section~\ref{ssec:baseline}, we consider keyword matching as simple baselines. 
The matching relies on a pre-defined speculative keywords dictionary. 
% A tuple is classified as speculative if its source sentences contain any word in the dictionary. 
A tuple is classified as speculative if its immediate neighbours in dependency parsing tree contain any word in the dictionary. 
We first use the 6 auxiliary modal verbs (`\textit{might}', `\textit{can}', `\textit{will}', `\textit{would}', `\textit{should}', and `\textit{had}') as the dictionary.
Shown in Table~\ref{tab:baseline_systems}, modal verbs based matching leads to very low recall, as expected.
We then include more speculative keywords in our keywords dictionary.

Specifically, we sum the frequency of all frequent cues across four datasets in Table~\ref{tab:spec-cue}, and keep the 30 most frequent cues as exemplified in Table~\ref{tab:spec-cue-top}.
We then build three variants on top of the basic dictionary, by adding the top 10, 20, and 30 speculative cues to the dictionary.
The enriched dictionary largely increases the recall scores (see Table~\ref{tab:baseline_systems}).
We notice that including top 10 keywords significantly increases the recall of speculative class by 12.3\%, and recall of hard case speculations by 20.9\%. 
In comparison, including additional top 11-20 keywords only marginally increases the recall of speculative class by 0.4\%, and recall of hard case speculations by 0.8\%.
Furthermore, the increase of including additional top 21-30 keywords is negligible.

% In comparison, including additional top 11-20, and 21-30 keywords only marginally increases the recall by 0.4\% and 0.1\%, respectively.

\begin{table}[t]
    \centering
    \resizebox{\linewidth}{!}{%
    \begin{tabular}{ll|ll|ll}
    \toprule
     \multicolumn{2}{c|}{Top 1-10}  & \multicolumn{2}{c|}{Top 11-20} & \multicolumn{2}{c}{Top 21-30} \\ 
             Cue & Freq. & Cue & Freq. & Cue & Freq. \\
     \midrule
      may  & 1665 & appear & 260 & probable & 122 \\
    if & 1294 & might & 256 & probably & 121 \\
    or & 1017 & consider & 250 & study & 101\\
    suggest & 928 & whether & 247 & allege & 89\\
    can & 882 & seem & 228 & determine & 87 \\
    would & 832 & investigate & 221 & putative & 80\\
    could & 495 & likely & 221 & hypothesis & 77\\
    indicate & 404 & should & 213 & expect & 75 \\
    think & 362 & believe & 198 & will & 60\\
    possible & 304 & examine & 183 & regard & 58 \\
    \bottomrule
    \end{tabular}}
    \caption{The 30 most frequent speculation cues by summing the frequency in Table~\ref{tab:spec-cue}.}
    \label{tab:spec-cue-top}
\end{table}

\end{document}